\documentclass[10pt,twocolumn,letterpaper]{article}

\usepackage[pagenumbers]{cvpr} 

\usepackage{graphicx}
\usepackage{amsmath}
\usepackage{amssymb}
\usepackage{booktabs}
\usepackage{lipsum}

\usepackage[pagebackref,breaklinks,colorlinks]{hyperref}

\usepackage[capitalize]{cleveref}
\crefname{section}{Sec.}{Secs.}
\Crefname{section}{Section}{Sections}
\Crefname{table}{Table}{Tables}
\crefname{table}{Tab.}{Tabs.}

\begin{document}
\title{Agriculture-Vision Challenge 2022 -- The Runner-Up Solution for Agricultural Pattern Recognition via Transformer-based Models}

\author{
Zhicheng Yang\\
PAII Inc.\\
Palo Alto, CA, USA\\
{\tt\small zcyangpingan@gmail.com}
\and
Jui-Hsin Lai\\
PAII Inc.\\
Palo Alto, CA, USA\\
{\tt\small juihsin.lai@gmail.com}
\and
Jun Zhou\\
Ping An Bank Co., Ltd\\
Shenzhen, Guangdong, China\\
{\tt\small zhoujun521@pingan.com.cn}
\and
Hang Zhou\\
PAII Inc.\\
Palo Alto, CA, USA\\
{\tt\small joeyzhou1984@gmail.com}
\and
Chen Du\\
PAII Inc.\\
Palo Alto, CA, USA\\
{\tt\small chendu.future@gmail.com}
\and
Zhongcheng Lai\\
Ping An Bank Co., Ltd\\
Shenzhen, Guangdong, China\\
{\tt\small laizhongcheng748@pingan.com.cn}
}

\maketitle

\begin{abstract}
   The Agriculture-Vision Challenge in CVPR is one of the most famous and competitive challenges for global researchers to break the boundary between computer vision and agriculture sectors, aiming at agricultural pattern recognition from aerial images. In this paper, we propose our solution to the third Agriculture-Vision Challenge in CVPR 2022. We leverage a data pre-processing scheme and several Transformer-based models as well as data augmentation techniques to achieve a mIoU of 0.582, accomplishing the 2nd place in this challenge.
\end{abstract}

\section{Introduction}
\label{sec:intro}

Computer vision applications in agricultural domain has become one of hot topics nowadays, especially using remote sensing satellite images and aerial images. With the rapid development of deep learning methods, numerous research studies have proposed pioneer and practical solutions to various computer vision problems in agriculture \cite{huang2022unsupervised,zhong2019deep,qiao2019generative,kamilaris2018deep}. Aside from fruitful research achievements, various algorithm challenges have been held at top-tier conferences for global researchers in recent years, in order to explore more effective algorithms to solve the specific problems. The \emph{Agriculture-Vision Challenge} in CVPR since 2020 is one of most famous and competitive challenges in this interdisciplinarity study. It aims at applying computer vision algorithms to agricultural pattern recognition from high-resolution aerial images. This year, CVPR 2022 holds the 3rd Agriculture-Vision Challenge, and we form our team ``PAII-RS'' to participate in this contest.
\section{Materials and Methods}
\label{sec:model}

In this section, we elaborate on the given datasets, the pre-processing method, the proposed deep learning-based framework, and the test-time augmentation (TTA) strategy.

\subsection{Description of Dataset}

\begin{table*}
	\centering
	\caption{Information of the given and resampled datasets for training and validation categories.}
	\vspace{-4pt}
	\begin{tabular}{clcc}  
		\toprule
		\textbf{Class Index} & \textbf{Class Name} & \textbf{Original Amount (Train/Val)} & \textbf{Resampled Amount (Train/Val)} \\
		\midrule
		0 & Background & 56944 / 18334 & 75121 / 13642  \\
		1 & Double Plant & 6234 / 2322 & 10961 / 2294  \\
		2 & Drydown & 16806 / 5800 & 19320 / 3383  \\
		3 & Endrow  & 4481 / 1755 & 8544 / 1858  \\
		4 & Nutrient Deficiency & 13308 / 3883 & 14859 / 2610  \\
		5 & Planter Skip & 2599 / 1197 & 5361 / 1015  \\
		6 & Water & 2155 / 987 & 4132 / 721  \\
		7 & Waterway & 3899 / 696 & 6024 / 1109  \\
		8 & Weed Cluster & 11111 / 2834 & 14423 / 2773  \\
		\bottomrule
	\end{tabular}
	\label{tab:imagecount}
	\vspace{-4pt}
\end{table*}

\begin{figure*}[t]
	\centering
	\includegraphics[width=0.9\linewidth]{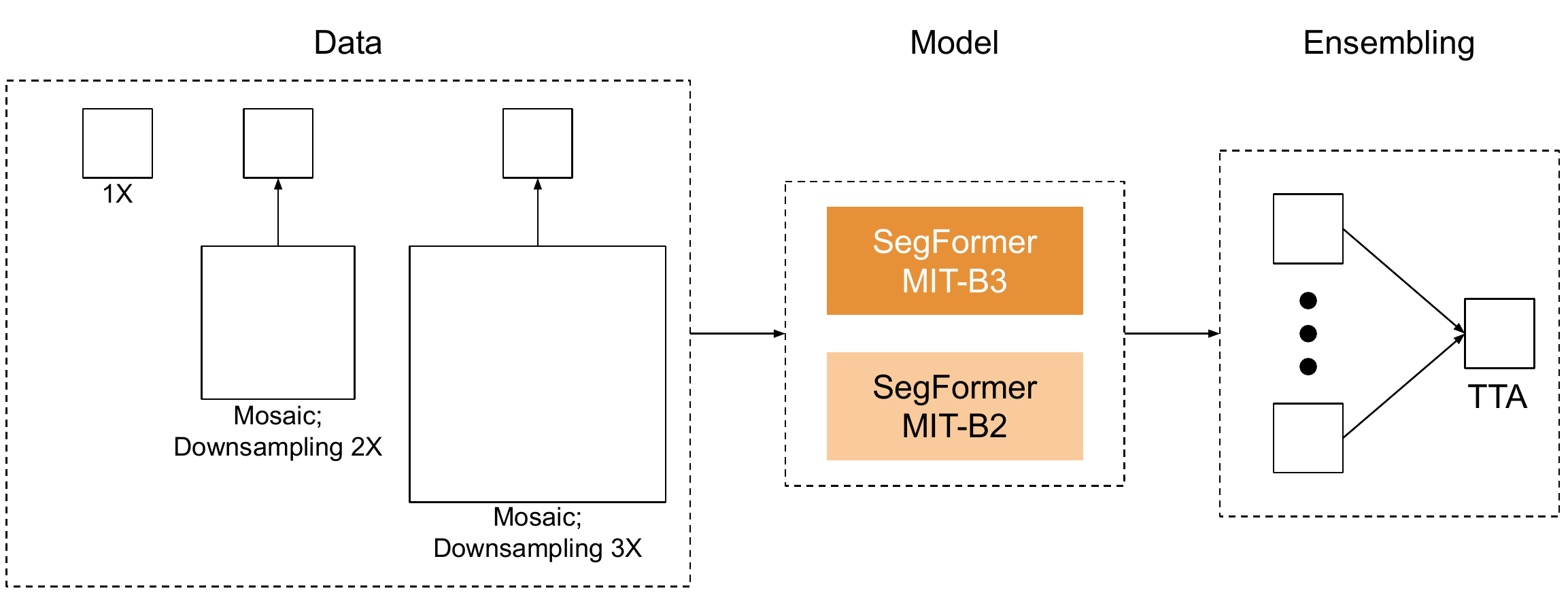}
	\vspace{-3pt}
	\caption{Framework of our deep learning-based model.}
	\label{fig:framework}
	\vspace{-3pt}
\end{figure*}

The challenge this year provides the entire Agriculture-Vision dataset released in \cite{chiu2020agriculture}. It contains 94,986 aerial farmland images collected throughout 2019 across the U.S. Each image has a size of 512$\times$512 pixels and has 4 channels (RGB and NIR). A total of 9 label classes are manually labeled for every image. Table~\ref{tab:imagecount} shows the given amount of images in each class. Note that many images have multiple labels, and even have overlapped labels (one pixel has multiple labels).

Although the amount of the given training data is considerable, we still generate more data following the data augmentation scheme of the winner solution last year. They conducted an image mosaic scheme to enable the model to have multi-scale views during the training. To fit the model input size, we create two new datasets using mosaicked images with down-sampling 2X (2 times) and down-sampling 3X as shown in Fig.~\ref{fig:framework}. The down-sampling dataset has the same image size of 512$\times$512 pixels that the recognition model can share the same network architecture among 1X, 2X, and 3X imagery.

\subsection{Data Pre-Processing}

We observe that the image counts in each category are uneven. For example, the image count of the background class is 25 times larger than the water class. To tackle the unbalance issue, we try to sample more images in the few-shot classes. The re-sampled image counts are listed in Table~\ref{tab:imagecount}. 

\subsection{Framework}

Fig.~\ref{fig:framework} shows our deep learning-based framework. \emph{SegFormer} is a Transformer-based efficient segmentation model \cite{xie2021segformer}. It designs a hierarchical Transformer encoder with multi-level feature outputs. Unlike other cumbersome decoders, SegFormer's decoder adopts MLP layers to aggregate multi-scale feature outputs from different layers. One of the key advantages of SegFormer is that its model size is relatively small but the performance keeps outstanding. Therefore, SegFormer is suitable for this challenge due to the model size parameter limit of 150M.

SegFormer provides six versions with various settings of Transformer encoders, leading to different model sizes. These six models are named from B0 to B5, with the increased model size. To follow the policy, we select Mix Transformer (MiT) B3 and Mix Transformer B2 as our training models. Their model size information can be found in Table~7 ``Mix Transformer Encoder'' in \cite{xie2021segformer}. After obtaining the individual inference result from each model, the model ensemble is performed to predict the final segmentation results.

\subsection{Test-Time Augmentation}
Since our models are trained with 1X, 2X, and 3X down-sampling imagery, we conduct the same processing on the test dataset. In addition to the scale augmentation, we include image rotation and flip.

\section{Results}

\subsection{Evaluation Metric}
The required evaluation metric is the average Intersection over Union metric (mIoU), which is defined as Eq.~\ref{eq:miou} to measure the performance.

\begin{equation}
	\text{mIoU} =\frac{1}{c} \sum \frac{\text { Area }\left(P_{c} \cap T_{c}\right)}{\text { Area }\left(P_{c} \cup T_{c}\right)}
	\label{eq:miou}
\end{equation}
where $c$ is the number of label classes (8 foreground classes + 1 background class for this challenge); $P_c$ and $T_c$ are the predicted label mask and ground truth label mask of the class $c$, respectively.

\begin{table*}
	\small
	\centering
	\caption{Performance comparisons among various models. The bold font of numeric results indicates the best performance on the test set. BG: Background; DP: Double Plant; D: Drydown; E: Endrow; ND: Nutrient Deficiency; PS: Planter Skip; W: Water; WW: Waterway; WC: Weed Cluster. The number in the parentheses following the class name refers to the class index.}
	\begin{tabular}{lcccccccccc}  
		\toprule
		\textbf{Models} & \textbf{mIoU} & BG(0) & DP(1) & D(2) & E(3) & ND(4) & PS(5) & W(6) & WW(7) & WC(8) \\
		\midrule
		\textit{(Other methods, on the \textbf{val} set)} & &  &  &   &  & &  &  & & \\
		Agriculture-Vision baseline(RGBN) \cite{chiu2020agriculture}  & 0.434 & 0.743 & 0.285 & 0.574 & 0.217 & 0.389 & 0.336 & 0.736 & 0.344 & 0.283 \\
		MiT-B3(RGBN)\cite{shen2022aaformer} & 0.454 & 0.768 & 0.371 & 0.609 & 0.245 & 0.424 & 0.413 & 0.692 & 0.269 & 0.299 \\
		MiT-B5(RGB)\cite{tavera2022augmentation} & 0.464 & 0.755 & 0.370 & 0.585 & 0.227 & 0.313 & 0.414 & 0.802 & 0.401 & 0.304 \\
		MiT-B5(RGBN)\cite{tavera2022augmentation} & 0.490 & 0.762 & 0.373 & 0.618 & 0.246 & 0.428 & 0.420 & 0.813 & 0.437 & 0.318 \\
		\midrule
		\textit{(Our implementation, on the \textbf{test} set)} & &  &  &   &  & &  &  & &\\
		HRNet-W48+OCR\cite{yuan2020object}(RGB baseline)  & 0.413 & 0.717 & 0.316 & 0.567 & 0.233 & 0.269 & 0.283 & 0.718 & 0.289 & 0.326 \\
		MiT-B3\cite{xie2021segformer}(RGB baseline)  & 0.448 & 0.720 & 0.395 & 0.557 & 0.325 & 0.364 & 0.330 & 0.687 & 0.293 & 0.358 \\
		MiT-B2\cite{xie2021segformer}(RGBN+Our method) & 0.554 & \textbf{0.778} & 0.483 & 0.632 & 0.476 & 0.570 & 0.403 & 0.768 & 0.410 & 0.466 \\
		MiT-B3\cite{xie2021segformer}(RGBN+Our method) & 0.563 & 0.773 & 0.471 & 0.640 & 0.452 & 0.569 & 0.442 & \textbf{0.782} & 0.463 & 0.475 \\
		\midrule
		\textit{(Our implementation, on the \textbf{test} set)} & &  &  &   &  & &  &  & &\\
		Model Ensemble(RGBN+Our method) & \textbf{0.582} & 0.777 & \textbf{0.485} & \textbf{0.646} & \textbf{0.481} & \textbf{0.573} & \textbf{0.471} & 0.779 & \textbf{0.547} & \textbf{0.479} \\
		\bottomrule
	\end{tabular}
	\label{tab:result}
\end{table*}

\subsection{Experiment Results}
Table~\ref{tab:result} presents our results, the baseline provided by the host Agriculture-Vision organizers, and the results of other methods. Note that other baselines evaluate their performance on the validation set due to the unavailable test set. As we can see, while our single model baselines are competitive with other baselines, our proposed method effectively improves the single model performance. Even though some single models have peak performance in some classes (0.778 for ``Background'' and 0.782 for ``Water''), our model ensemble enjoys the merits of multiple single models' strength to achieve the mIoU of 0.582. It also shows that our ensemble results significantly outperform other baselines and our implementation of various single models. 
\section{Conclusions}
\label{sec:concl}

In this paper, we propose our solution to the 3rd Agriculture-Vision Challenge in CVPR 2022. For data usage, we perform data pre-processing and test data augmentation schemes. Several SegFormer models are leveraged. We finally accomplish a mIoU of 0.582, achieving the 2nd place in this challenge. 

\textbf{Future Directions.} The potential applications of our proposed algorithm include crop type identification in precision agriculture, agricultural asset estimation and agricultural insurance product design in the Environmental, Social, and Governance (ESG) domain. These future directions can illuminate the revitalization of rural areas and facilitate the service of inclusive finance in an eco-friendly way.

{\small
\bibliographystyle{ieee_fullname}
\bibliography{MAIN}
}

\end{document}